\documentclass[10pt,twocolumn,letterpaper]{article}
\usepackage{wacv}              
\usepackage{multirow}
\usepackage{amsmath}
\usepackage{pifont}
\usepackage{float}
\usepackage{graphicx}    
\usepackage{caption} 
\usepackage{dblfloatfix} 
\usepackage{booktabs}
\usepackage[english]{babel}
\usepackage{microtype}
\usepackage{subcaption} 
\usepackage{colortbl}     
\usepackage[utf8]{inputenc} 
\usepackage{kotex}  
\usepackage{cuted} 
\newcommand{\Sref}[1]{Sec.~\ref{#1}}

\newcommand{\Tref}[1]{Table~\ref{#1}}
\usepackage{xspace}
\def\onedot{.\@\xspace}
\def\eg{\emph{e.g}\onedot} 
\def\ie{\emph{i.e}\onedot}

\definecolor{yw}{rgb}{0.01176, 0.5490, 0.5490}

\usepackage{etoolbox}
\usepackage{mdframed} 
\definecolor{wacvblue}{rgb}{0.21,0.49,0.74}
\usepackage[pagebackref,breaklinks,colorlinks,allcolors=wacvblue]{hyperref}

\def\wacvPaperID{1463} 
\def\confName{WACV}
\def\confYear{2026}

\title{mEOL: Training-Free Instruction-Guided Multimodal Embedder \\for Vector Graphics and Image Retrieval}

\author{
Kyeong Seon Kim\textsuperscript{1} \quad
Baek Seong-Eun\textsuperscript{2} \quad
Lee Jung-Mok\textsuperscript{2} \quad
Tae-Hyun Oh\textsuperscript{1*}
\\[4pt]
\textsuperscript{1}KAIST \qquad
\textsuperscript{2}POSTECH
\\[4pt]
\texttt{\{ellakim,taehyun.oh\}@kaist.ac.kr}, \quad
\texttt{\{seongeun,jungmok\}@postech.ac.kr}
} 
\vspace{-10em}

\begin{document}
\twocolumn[{%
\vspace{-1.0em}
  \renewcommand\twocolumn[1][]{#1}%
  \maketitle
  \begin{center}
    \includegraphics[width=\textwidth]{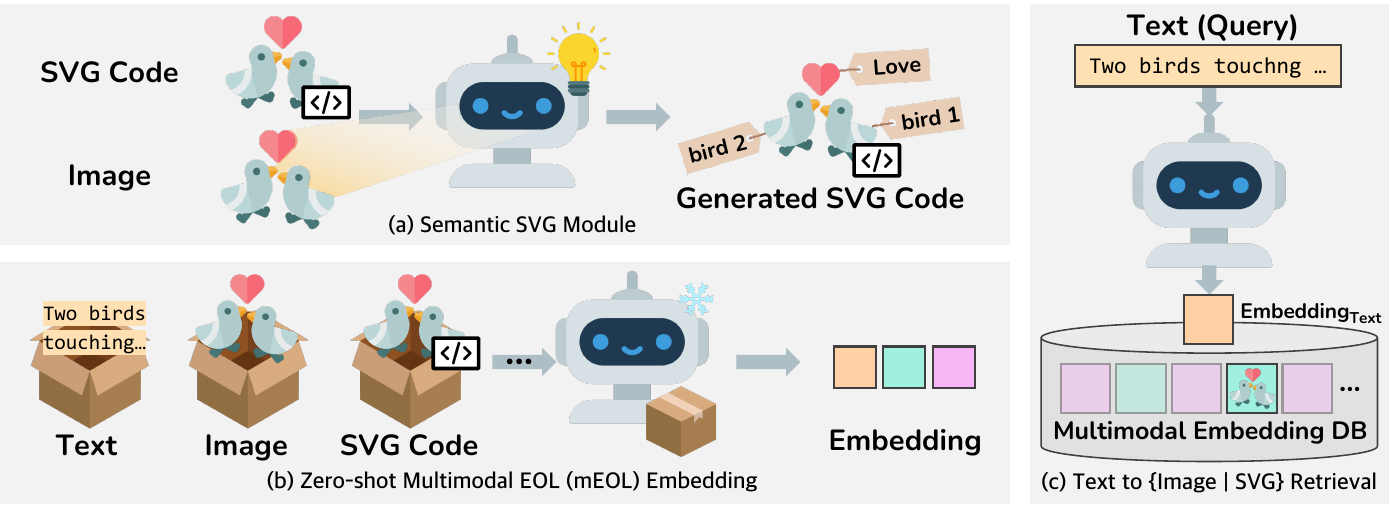}
    \captionof{figure}{ 
    \textbf{Overview of Our Training-free Multimodal Embedding Method.}
    (a) The semantic SVG module rewrites raw SVG code by assigning meaningful identifiers and simplifying structure through MLLM-based visual reasoning.
(b) Multimodal Explicit One-word Limitation (mEOL) prompts instruct the MLLM to summarize text, images, and SVG code into a single-token embedding, placing all modalities in an aligned space.
(c) The resulting embeddings enable cross-modal retrieval, allowing natural-language queries to retrieve relevant images and SVGs without any training.
    }
    \label{fig:teaser}
  \end{center} 
}]
\begin{abstract}

Scalable Vector Graphics (SVGs) function both as visual images and as structured code that encode rich geometric and layout information, yet most methods rasterize them and discard this symbolic organization. At the same time, recent sentence embedding methods produce strong text representations but do not naturally extend to visual or structured modalities. 
We propose a training‑free, instruction‑guided multimodal embedding framework that uses a Multimodal Large Language Model (MLLM) to map text, raster images, and SVG code into an aligned embedding space. We control the direction of embeddings through modality‑specific instructions and structural SVG cues, eliminating the need for learned projection heads or contrastive training. 
Our method has two key components: (1) Multimodal Explicit One-word Limitation (mEOL), which instructs the MLLM to summarize any multimodal input into a single token whose hidden state serves as a compact semantic embedding. 
(2) A semantic SVG rewriting module that assigns meaningful identifiers and simplifies nested SVG elements through visual reasoning over the rendered image, exposing geometric and relational cues hidden in raw code.
Using a repurposed VGBench, we build the first text‑to‑SVG retrieval benchmark and show that our training‑free embeddings outperform encoder‑based and training‑based multimodal baselines. These results highlight prompt‑level control as an effective alternative to parameter‑level training for structure‑aware multimodal retrieval. Project page: \href{https://scene-the-ella.github.io/meol/}{https://scene-the-ella.github.io/meol/}
\end{abstract}

\vspace{-2.5em}
\section{Introduction}
Scalable Vector Graphics (SVGs)~\cite{ferraiolo2000scalable} are widely used in user interfaces~\cite{carlier2020deepsvg} because they are compact, resolution-independent, and encode both visual content and structural information.
They use geometric attributes to define shapes and optional identifiers to group elements or suggest semantic roles.
For instance, a circle might be defined by \texttt{<circle cx="50" cy="50" r="40"/>} for geometry, while related elements can be grouped under an identifier \texttt{<g id="icon-group">}. 
These structural components encode more than just appearance. They can express relationships between parts, object categories, or semantic intent. For example, grouping elements under \texttt{<id="love\_bird">} or \texttt{<id="wing">} introduces meaning that goes beyond raw geometry.
However, many existing methods convert SVGs into raster images before analysis, and discard structural cues and treating SVGs as only visual inputs.
As a result, downstream tasks such as retrieval, question answering, or semantic search suffer from degraded performance due to the loss of high-level structural and symbolic information~\cite{wang2024visually}.
To address these limitations, we propose a training-free multimodal embedding method that preserves and utilizes the structural information in SVGs. 
Our method embeds text, raster images, and SVG code into an aligned embedding space using a Multimodal Large Language Model (MLLM).
The method consists of two components: First, we propose a training-free multimodal embedder using Explicit One-word Limitation (mEOL), which compresses diverse inputs into a single-word embedding. 
Second, we propose a semantic SVG module that rewrites SVG code by generating meaningful identifiers and simplifying structure.
These allows the model to leverage both appearance and symbolic structure for cross-modal retrieval, without any training or fine-tuning.
\noindent Our main contributions are summarized as follows:
\begin{itemize}
    \item \textbf{Instruction-guided multimodal embeddings.}
    We propose mEOL, a training-free instruction-guided multimodal embedder using an MLLM that supports retrieval across text, raster image, and SVG code modalities. 
    Even when given long-form natural language queries, our method can retrieve semantically relevant results using both appearance and structure. 
    \item \textbf{Semantic SVG rewriting for structure-aware embeddings.}
    We introduce a semantic SVG module that rewrites raw SVG code into a more interpretable symbolic program by assigning meaningful identifiers to grouped elements and simplifying nested structures, guided by visual reasoning. This module exposes geometric and relational cues that are invisible to raster-only processing while preserving the original image appearance, making SVG structure usable for retrieval.
    \item \textbf{Benchmarking and analysis for text-to-SVG and image retrieval.}
    We repurpose VGBench to construct, to our knowledge, the first text-to-SVG retrieval benchmark, and evaluate text-to-image and text-to-image+SVG retrieval.
\end{itemize}
\section{Related Work}
\begin{figure*}[!t]  
  \centering
  \begin{subfigure}[t]{0.64\textwidth}
    \centering
    \includegraphics[width=\linewidth]{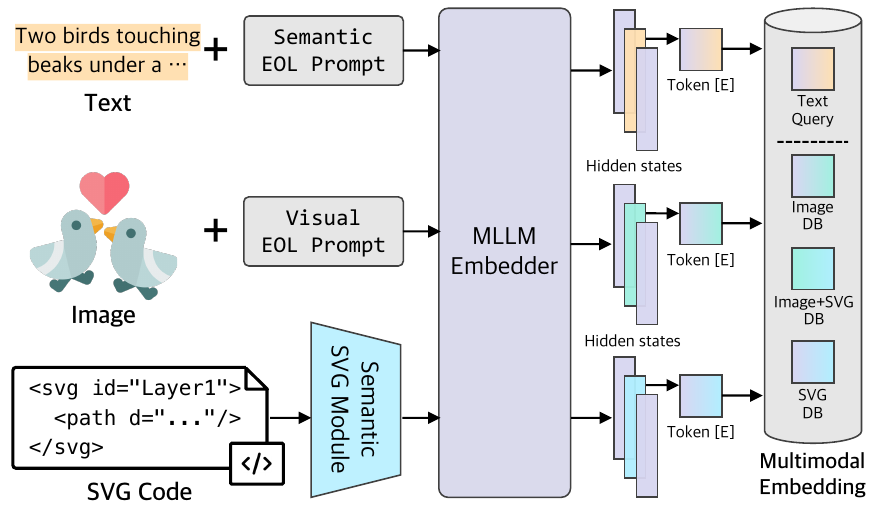}
    \caption{Illustration of our training-free multimodal EOL embedding.} 
    \label{fig:onewordtoken}
  \end{subfigure}
  \hfill
  \begin{subfigure}[t]{0.35\textwidth}
    \centering
    \includegraphics[width=\linewidth]{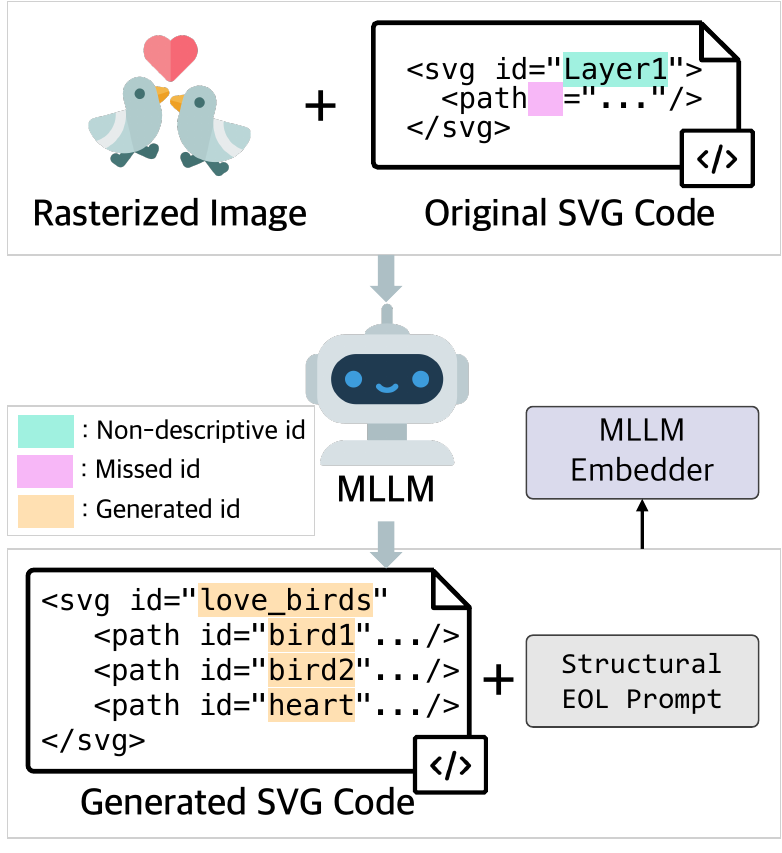}
    \caption{Pipeline of the semantic SVG module.}
    \label{fig:enhancement}
  \end{subfigure}
  \caption{\textbf{Components of Our Training-free Multimodal Embedding pipeline.} \textmd{
  (a) Given text, image, or SVG inputs, each input is paired with a multimodal EOL (mEOL): Semantic EOL Prompt for text and Visual EOL Prompt for images.
  Then, our training-free MLLM embedder generates an aligned embedding.
  The embedding is extracted from the hidden state of the final output token \texttt{[E]} at the penultimate layer.
  (b) Before SVG embedding, the Semantic SVG Module generates SVGs by completing missing or non-descriptive ids (\eg, \texttt{id="Layer\_1"} to \texttt{id="bird1"}) and simplifying the code structure through visual reasoning over the rendered image. 
  This process improves the structural grounding of embeddings without altering visual appearance.
  The Generated SVG with the Structural EOL are then fed into the MLLM embedder to produce SVG or Image+SVG embeddings.
  }}
  \label{fig:combined}
\end{figure*}
\subsection{Scalable Vector Graphics (SVGs)}
Scalable Vector Graphics (SVGs) encode visual information through geometric paths and structural attributes representing shapes and paths~\cite{cai2025investigation}.
These elements not only make them well-suited for scalable applications~\cite{xing2024svgfusion, tang2024strokenuwa, bing2024deepicon, hu2024vectorpainter, yang2025omnisvg} across various devices and resolutions, but also convey more than appearance.
However, previous SVG studies such as 
generation~\cite{jain2023vectorfusion, xing2024svgdreamer, song2025layertracer, wu2024chat2svg, rodriguez2023starvector}, understanding~\cite{zou2024vgbench,chen2025svgenius, cai2025investigation, qiu2408can} and SVG-to-SVG retrieval~\cite{cao2023svgformer} 
remain challenging because lengthy and complex structures are difficult for LLMs or MLLMs to understand without additional training~\cite{cai2023leveraging, wang2024visually}.
To address this challenge, recent studies have transformed complex SVGs into alternative representations
to help models better grasp the content and context of SVG code~\cite{wang2024visually,lopes2019learned} and improve SVG-to-SVG retrieval performance~\cite{cao2023svgformer}.
However, such methods generally rely on additional model fine-tuning and
these approaches do not support text-to-SVG retrieval. In other words, retrieval over structured SVGs from natural language remains largely unexplored, especially in training-free settings. Therefore, we propose the first training-free method designed specifically for text-to-SVG retrieval tasks, which addresses this gap by regenerating SVGs and injecting semantics to make them interpretable by MLLMs.

\subsection{Sentence Embedding}
Sentence embedding maps a sentence into a compact vector that captures its semantic meaning. 
Early approaches used pretrained encoders like BERT~\cite{jiang2022promptbert, reimers2019sentence}, or contrastive learning methods~\cite{gao2021simcse, chuang2022diffcse}, which train the model to bring semantically similar sentences closer in embedding space.
More recent methods leverage large language models (LLMs) for embedding~\cite{lee2024nv, behnamghader2024llm2vec, meng2024sfrembedding, ni2021sentence, muennighoff2024generative}. These involve fine-tuning the model or training additional layers for specific downstream tasks.
In contrast, some methods extract sentence embeddings without any parameter updates by guiding the LLM to compress a sentence into a single word.
This strategy, known as Explicit One-word Limitation (EOL), prompts the model with a template such as \texttt{"This sentence: [text] means [MASK]"}.
Variants such as PromptEOL~\cite{jiang2024scaling}, KEEOL~\cite{zhang2024simple}, MetaEOL~\cite{lei2024meta}, and GenEOL~\cite{thirukovalluru2024geneol} differ mainly in how they modify the prompt. These changes alone have shown strong performance improvements in training-free setups.
However, existing EOL methods are limited to plain text. They do not generalize to non-text modalities such as images or structured SVG code.
We are the first to extend EOL to multimodal inputs.
\subsection{Multimodal Joint Embedding}
Embedding is a fixed-size vector representation that contains the semantics of input data and can be used for diverse downstream tasks such as classification, clustering, reranking, and retrieval tasks~\cite{muennighoff2022mteb, jiang2022promptbert, reimers2019sentence, gao2021simcse, chuang2022diffcse, choi2026PatchwiseRetrieval, bang2026bth}. 
Multimodal embedding methods aim to align representations across different modalities, such as text and image. 
Earlier models such as ~\cite{clip, li2022blip, zhai2023sigmoid, NEURIPS2021_50525975} use modality-specific encoders to project modalities and train them with contrastive learning on large paired datasets.
More recent approaches leverage MLLMs~\cite{jiang2024vlm2vec, lan2025llave}, which enable to encode both images and text. 
Such methods often train the model with contrastive losses to effectively produce aligned embeddings.
However, such training-based approaches can be costly, especially for specialized data formats (\eg, SVG). 
Therefore, we propose a training-free multimodal embedding method for SVG using MLLMs. By extending EOL to non-text modalities, our approach enables aligned embeddings across text, raster images, and SVG code, without any fine-tuning.

\section{Methodology}
In this section, we propose a semantic SVG module (\Sref{sec:svg-enhancement}) and a zero-shot multimodal Explicit One-word Limitation (mEOL) embedder
(\Sref{sec:Zero-shot Multimodal Embedder}).

\subsection{Pipeline Overview}
Our pipeline consists of three main steps, as shown in Fig.~\ref{fig:combined}. (1) Generating SVGs using Semantic SVG module, (2) Training-free multimodal embedding, and (3) Retrieval using cosine similarity. First, we regenerate the SVG code by replacing non-descriptive identifiers with meaningful labels (\eg, \texttt{"Layer\_1"} to \texttt{"bird"}) and simplifying the code using a MLLM, which enriches the semantic information without changing visual appearance. Next, the generated SVG codes, images, and textual inputs are fed into the MLLM with each prompt, generating embeddings represented by the hidden state of the last token that compactly represent both visual and semantic features. Finally, retrieval is performed by computing cosine similarity between query and database embeddings in an aligned representation space. 
Through these steps, our pipeline can address text-to-SVG retrieval effectively, bridging the gap between various modalities.

\subsection{Semantic SVG Module}
\label{sec:svg-enhancement}
SVG code often includes attributes such as \texttt{id} and \texttt{class} that are useful for styling or scripting, but many SVGs contain non-descriptive or missing identifiers (\eg, \texttt{id="Layer\_1"}, \texttt{id="path123"}). 
In addition, SVGs frequently contain overly complex and redundant structures, making them difficult for language models to interpret.
As a result, existing methods typically convert SVGs into raster images, discarding their symbolic structure and limiting the possibility.

To address this, we propose a Semantic SVG Module that completes and leverages the structural information of SVG code.
The module jointly analyzes the raw SVG code and its rendered image using MLLM, and generates a semantically enriched and compact version of the input.

The overall process of the semantic SVG module (Fig.~\ref{fig:enhancement}) consists of four components: 
(1) Input: The MLLM takes the original SVG code along with its rendered image. 
(2) Image \& SVG Analysis: It analyzes both the SVG structure and its rendered image to identify visually salient objects (\eg, \texttt{"bird"}) by using the joint context of code and image. 
(3) Object Matching \& Code Simplification: Detected visual objects are aligned with the corresponding SVG elements, and redundant or unnecessarily nested components are simplified based on geometric structure and relative position in the code and image. 
(4) ID assignment: Each matched SVG element is assigned a meaningful identifier based on the detected object, replacing non-descriptive labels (\eg, \texttt{id="Layer\_1"}) with semantic ones (\eg, \texttt{id="bird"}), or assigning new IDs to elements with missing identifiers.
Generated SVG encodes both symbolic and structural information that was previously ignored. 
This enables, for the first time, training-free retrieval from text to structured SVG code. Combined with the SVG EOL Prompt, this generated SVG code significantly improves embedding quality and retrieval performance.
\subsection{Training-free Multimodal Embedder}
\label{sec:Zero-shot Multimodal Embedder}
In this section, we propose a novel training-free multimodal embedding method, extending Explicit One-word Limitation (EOL)~\cite{zhang2024simple, thirukovalluru2024geneol} to support diverse modalities, including text, raster images, and SVG code. We refer to this extension as \textit{multimodal EOL (mEOL)}.
Our method guides a MLLM to generate a compact semantic representation for each modality using mEOL to summarize the input as a single word, as shown in Fig.~\ref{fig:onewordtoken}. 
MLLMs operate autoregressively, generating one token at a time based on prior context.
Using this property, we prompt the model with a concise instruction \texttt{"$[X]\ \text{means in one word:} $"}.
Given a multimodal input $[X]$, the model generates a single token that summarizes the input’s semantic meaning. 
Each modality's instruction encourages the model to focus on the most salient features of each modality and to extract a condensed semantic representation such as structural or visual cues as a single output token:
\begin{itemize}
    \item \textbf{Text.} The instruction emphasizes both the semantic meaning of the description and its connection to visual attributes such as shape or layout.
    \item \textbf{Image.} The instruction guides the model to focus on visual features including color, structure, and object-level details.
    \item \textbf{SVG.} The instruction requests interpretation of the vector code in terms of its rendered meaning, directing attention to symbolic groupings and visual structure.
    \item \textbf{Image+SVG}. The model is instructed to jointly reason over rasterized appearance and SVG semantics, capturing both rendered style and underlying structure.
\end{itemize}
To generalize and support inputs from different modalities, our mEOL designs distinct modality-aware instructions tailored to each input type:
\vspace{2pt}
\boxed{\texttt{This $\langle \text{[X]} \rangle$ $\langle \text{instruction} \rangle$ in one word:}}
\vspace{2pt}
This format is consistently applied to text, images, and SVG code. The detailed instructions for each modality are provided in Supplementary Table 1. 
Conditioned on mEOL, we extract hidden states across the layers, denoted as $\mathcal{H} = [h^{(1)}, h^{(2)}, \ldots, h^{(L)}]\in\mathbf{R}^{L\times T}$, where $L$ is number of layers and $T$ is the whole token length. Among those hidden states, we select the hidden state of last token at the penultimate layer as the final embedding:
$$
e_{mEOL} = h^{(L-1)}_{\text{T}}
$$
This representation compactly captures the semantic summary of the input, informed by MLLM's rich prior knowledge.
Compared to the conventional feature extractors, our multimodal embedder first projects various modalities into the MLLM's aligned text space, and makes embeddings into one using MLLM's strong prior knowledge. 
By directing the model to ground modality-dependent information into one aligned representation anchor, we enable training-free retrieval across modalities, including structured SVG code and image which prior EOL methods could not support.
\section{Results and Analysis}
\begin{table*}[t]
    \centering
    \resizebox{0.9\linewidth}{!}{
    \begin{tabular}{c|c|cccccc}
    \toprule
    \midrule
    \multirow{2}{*}{Model} & \multirow{2}{*}{Training}    & \multicolumn{6}{c}{Query: Text / Database: Image} \\ \cmidrule{3-8}
    &                           & \multicolumn{1}{c}{Recall@1} & \multicolumn{1}{c}{Recall@5} & \multicolumn{1}{c}{Recall@10} & \multicolumn{1}{c}{Recall@15} &\multicolumn{1}{c}{Recall@20} & \multicolumn{1}{c}{MRR} \\ \midrule
    CLIP~\cite{clip}   &   \textcolor{red}{\ding{55}}   &  0.1453   &  0.2442  &  0.2733   &  0.2878   &  0.2951  &  0.1854  \\
    BLIP~\cite{li2022blip} & \textcolor{red}{\ding{55}}      &  0.2078   &  0.3081  &  0.3387   &  0.3547   &  0.3648  &  0.2521  \\
    SigLIP~\cite{zhai2023sigmoid} & \textcolor{red}{\ding{55}}     &  0.0465   &  0.1076  &  0.1483   &  0.1628   &  0.1788  &  0.0780  \\
    \midrule
    VLM2Vec~\cite{jiang2024vlm2vec} & \textcolor[HTML]{228B22}{\ding{51}}  &  0.2137   &  0.3110  &  0.3314   &  0.3459   &  0.3532  &  0.2567  \\
    LLaVE~\cite{lan2025llave}   & \textcolor[HTML]{228B22}{\ding{51}}       &  0.2326   &  0.3343  &  0.3561   &  0.3706   &  0.3735  &  0.2757  \\ 
    LLaMA-3.2-11B~\cite{grattafiori2024llama} (Ours)   & \textcolor{red}{\ding{55}} &  0.3387   &  0.5785  &  0.6526   &   0.6933  & 0.7224  &  0.4428 \\
    Qwen2.5-VL-7B~\cite{bai2025qwen2} (Ours) & \textcolor{red}{\ding{55}} & \textbf{0.3503}   &  \textbf{0.6177}  &  \textbf{0.6846}   &   \textbf{0.7137}  & \textbf{0.7456}  & \textbf{ 0.4634} \\ \midrule \bottomrule
    \end{tabular}
    }
    \caption{\textbf{Text-to-Raster Image Retrieval Performance on the Repurposed VGBench Dataset.} \textmd{We compare our method with commonly used Vision-Language Models (VLMs) in text-to-image retrieval~\cite{clip, li2022blip, zhai2023sigmoid} and embedding methods utilizing Multimodal Large Language Model (MLLM)~\cite{jiang2024vlm2vec, lan2025llave}. Our method outperforms prior VLMs and MLLM-based approaches across all top-k recall metrics.}} 
\label{tab:experiments1}
\end{table*}

\begin{figure*}[ht]
  \centering
  \includegraphics[width=\linewidth]{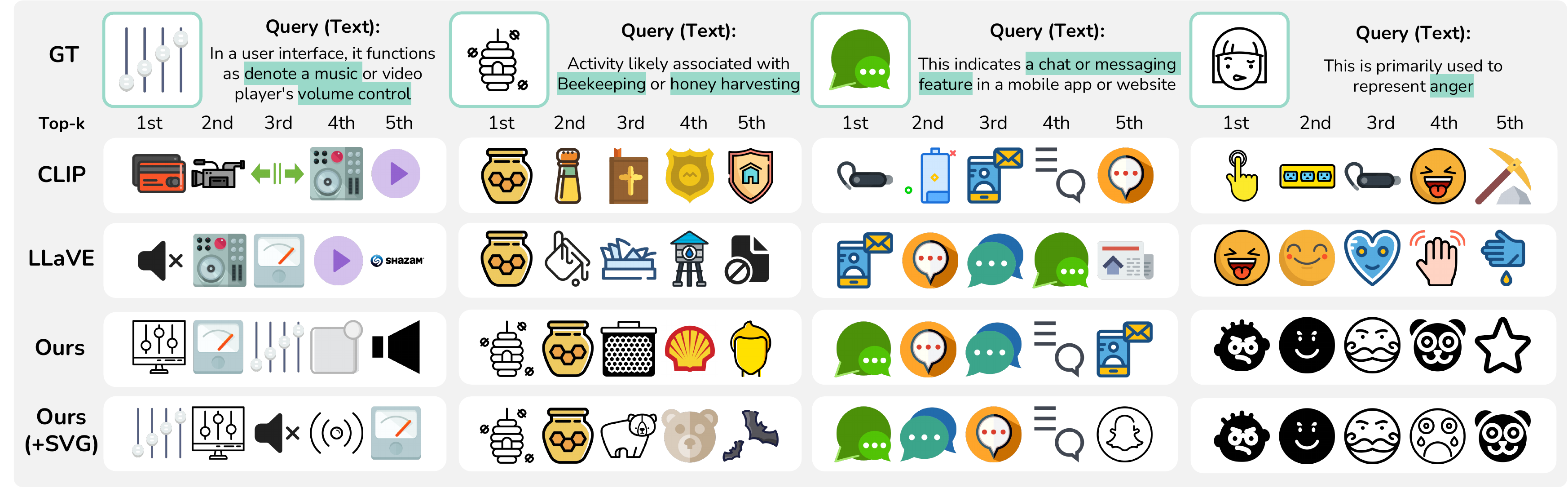}
  \caption{\textbf{Qualitative Examples of Text-to-Image, Text-to-Image with SVG Code Retrieval.} \textmd{The icon with the yellow border represents the target image. Given a text query, each model retrieves images ranked from 1st to 5th based on cosine similarity. Compared with CLIP and LLaVE, our methods retrieved more semantically aligned with the query.}}
  \label{fig:qualitative_result}
\end{figure*}
We first describe our experimental settings in \Sref{sec:4.1}. In \Sref{sec:4.2}, we demonstrate the effectiveness of our method by evaluating its performance and qualitative results in text-to-raster-image retrieval performance. In~\Sref{sec:4.3}, we extend our method to additional modalities and show that our method implements the most effective EOL method compared to existing EOL approaches. Finally, we present appropriate settings for extracting embeddings and analyses through various ablation studies in~\Sref{sec:4.4}.

\subsection{Experimental Settings}
\begin{center}
\renewcommand{\arraystretch}{1}
\setlength{\tabcolsep}{6pt}
\resizebox{\linewidth}{!}{
\begin{tabular}{lcccc}
\toprule
\textbf{Subset} & \textbf{Category} & \textbf{\# of Samples} & \textbf{Avg. SVG Length} & \textbf{Components} \\
\midrule
SVG~\cite{wang2024visually} & Usage & 1,426 & $\sim$2,400 & Q, option (A-D), A \\
\bottomrule
\end{tabular}
}
\captionof{table}{\textbf{Statistics of the Repurposed VGBench Used in Our Experiments}. We adapt VGBench dataset for the text-to-SVG retrieval task by repurposing its annotations.}
\label{tab:vgbench_stats}
\end{center}

\label{sec:4.1}
\label{sec:experiment_settings}
We evaluate our method using a repurposed version of the SVG dataset from VGBench~\cite{wang2024visually}, originally designed for vector graphic question answering (VGQA). 
The dataset contains thousands of instances covering unique SVG images. 
Each instance includes an SVG code, a natural language question, four multiple-choice options (A–D), and a ground-truth answer label. 
Among those, we specifically use the subset of Usage category for SVG question answering.
For example, a question such as \textit{"What does this SVG image likely represent?"} may be accompanied by options like \textit{[A: Global connectivity issues, B: Worldwide music distribution, C: Ecological awareness, D: Travel booking services]}, with the correct answer being \textit{A}. 

To reframe this into a text-to-visual retrieval task, we concatenate the question and correct answer A into a single natural language query (Q+A), evaluate whether the correct raster image can be retrieved from multimodal embedding database. 
It provides a controlled and reproducible setup for evaluating multimodal embedding quality.
The detailed statistics in Table~\ref{tab:vgbench_stats}. On average, SVGs in VGBench contain approximately 2,400 characters of raw XML code, reflecting nontrivial structural complexity. As illustrated in Fig.~\ref{fig:onewordtoken}, we treat this textual input as the \textbf{query}, while the retrieval \textbf{database} is composed of all 1426 examples, represented in three formats: (1) rasterized image, (2) raw SVG code, and (3) a combination of both. 
This setup allows us to evaluate retrieval across different modalities such as text-to-image, text-to-SVG, and text-to-hybrid (image+SVG) without any training. 

The retrieval database (DB) consists of SVG codes and raster images used throughout the VGQA Usage subset. 
For each query, we compute its embedding using our method and perform cosine similarity retrieval against the DB. We report standard retrieval metrics, including Recall@1, Recall@5, and Recall@10, and Mean Reciprocal Rank (MRR) to assess how well the model retrieves the ground-truth icon based on its description. Recall@k measures whether the correct item appears in the top-k results, while MRR reflects the inverse rank of the first correct retrieval result, and provides a measure of how early the relevant item appears in the ranked list.
For comparative baselines, we have selected the encoder-based methods ~\cite{clip, li2022blip, zhai2023sigmoid} and MLLM-based encoding methods~\cite{jiang2024vlm2vec, lan2025llave}. All models are evaluated under the same retrieval setup using cosine similarity between the extracted text and raster image embeddings.

For our training-free multimodal embedding method, we have used two MLLMs: LLaMA-3.2-11B and Qwen2.5-VL-7B. We used the last token hidden states of the penultimate layer as embeddings. 
We provide empirical evidence supporting our embedding design choices in~\Sref{sec:4.4}, focusing on embedding method and layer selection.
\begin{table*}[t]
    \resizebox{1\linewidth}{!}{
        \begin{tabular}{c|c|ccccccccc}
        \toprule \midrule
        \multirow{3}{*}{Model}      & \multirow{3}{*}{EOL} & \multicolumn{9}{c}{Database format (Query: Text)} \\ \cmidrule{3-11}  &   & \multicolumn{3}{c|}{SVG} & \multicolumn{3}{c|}{Image}  & \multicolumn{3}{c}{Image + SVG}  \\&                & Recall@1 & Recall@10 & \multicolumn{1}{c|}{Recall@20} & Recall@1 & Recall@10 & \multicolumn{1}{c|}{Recall@20} & Recall@1 & Recall@10 & \multicolumn{1}{c}{Recall@20} \\ \midrule
        \multirow{3}{*}{Mistral-7B}
        & PromptEOL  &  0.0087 &   0.0203   & \multicolumn{1}{c|}{0.0392}       &       -   &      -    & \multicolumn{1}{c|}{-}          &      -    &    -      & \multicolumn{1}{c}{-}          \\
        & KEEOL  &     0.0087   &    0.0247     & \multicolumn{1}{c|}{0.0451}   &    -      &     -     & \multicolumn{1}{c|}{-}          &       -   &     -     & \multicolumn{1}{c}{-}   \\
        & Ours      &   \textbf{0.1439}     &   \textbf{0.2922} & \multicolumn{1}{c|}{\textbf{0.3314}}   &   -  &    -   & \multicolumn{1}{c|}{-}   &    -      &    -      & \multicolumn{1}{c}{-}          \\ \midrule
        \multirow{3}{*}{Qwen2.5-VL-7B} 
        & PromptEOL   &    0.0131 &    0.0218   & \multicolumn{1}{c|}{0.0422} & 0.3183   & 0.6483      & \multicolumn{1}{c|}{0.6846} &  0.3154  &  0.6628   & \multicolumn{1}{c}{0.7137}          \\
        & KEEOL     &      0.0116   &     0.0218  & \multicolumn{1}{c|}{0.0422}          & 0.2703    &     0.5494   & \multicolumn{1}{c|}{0.6017}      &    0.2500       & 0.5567       & \multicolumn{1}{c}{0.6177}          \\
        & Ours      &    \textbf{0.1744}   &   \textbf{0.2994}    & \multicolumn{1}{c|}{\textbf{0.3314}}   &  \textbf{0.3503}       & \textbf{0.6846}    & \multicolumn{1}{c|}{\textbf{0.7456}}    &   \textbf{0.3765}   &   \textbf{0.7224}  & \multicolumn{1}{c}{\textbf{0.7776}}          \\ \midrule
        \multirow{3}{*}{LLaMA-3.2-11B}  
        & PromptEOL    &   0.0102  &  0.0177   & \multicolumn{1}{c|}{0.0320} &  \textbf{0.3387}   &  \textbf{0.6555}      & \multicolumn{1}{c|}{0.7151} &  \textbf{0.3270}  &  0.6483  &  \multicolumn{1}{c}{0.7093}          \\
        & KEEOL        &   0.0087  &  0.0161   & \multicolumn{1}{c|}{0.0276} &  0.2762   & 0.5509       & \multicolumn{1}{c|}{0.6308}   &   0.2573  &  0.5203 & \multicolumn{1}{c}{0.6134}          \\
        & Ours      &      \textbf{0.1599}  &  \textbf{0.3038}   & \multicolumn{1}{c|}{\textbf{0.3474}} &  \textbf{0.3387}   & 0.6526    & \multicolumn{1}{c|}{\textbf{0.7224}}      &   0.3169   &     \textbf{0.6628}  & \multicolumn{1}{c}{\textbf{0.7427}}    \\ \midrule \bottomrule
        \end{tabular}}
    \caption{\textbf{Comparison of EOL Family and Baselines across Different Models and Database Formats.} \textmd{Our multimodal EOL with semantic SVG module significantly improves retrieval accuracy. Note that since Mistral-7B does not support image modality, related cells are left blank.}}
    \label{tab:eol_ablation}
    \vspace{-0.5em}
\end{table*}
\subsection{Main Results}
\label{sec:4.2}
\paragraph{Quantitative Results.}
In ~\Tref{tab:experiments1}, we report the retrieval performance of the SVG dataset, given the concatenated text query (Q+A), and raster image as database. It shows that our method consistently outperforms encoder-based approaches such as CLIP~\cite{clip}, BLIP~\cite{li2022blip}, and SigLIP~\cite{zhai2023sigmoid}, which rely on contrastive learning. Notably, our training-free approach exhibits strong generalizability across various Vision-Language Models (VLMs), achieving better performance than encoder-based methods without the need for additional training. While SigLIP suffers from limitations due to its maximum token length, potentially degrading its performance on longer queries, our method effectively handle lengthy and complex inputs by summarizing them into a compact word embedding. Furthermore, in evaluations against VLM2Vec~\cite{jiang2024vlm2vec} and LLaVE~\cite{lan2025llave}, our approach achieves superior performance despite not being explicitly trained for embedding extraction like such methods, highlighting its efficacy and robustness.
\vspace{-1em}
\paragraph{Qualitative Results.}
Figure~\ref{fig:qualitative_result} represents qualitative results of the top-5 retrieved SVG icons given a text query. We compare an encoder-based method, \ie, CLIP, and a MLLM-based method, \ie, LLaVE, along with our proposed method. For visualization clarity, we simplify the textual queries in the figure.
As illustrated in the first column, our method is the only one that successfully captures the intended concept of ``anger'', retrieving a semantically meaningful icon. In contrast, LLaVE retrieves generic face-like icons, and CLIP fails to retrieve a relevant icon, retrieving only a loosely related result at the 4th position.
In the second column, our method demonstrates that the semantic SVG module contributes to capturing not only the structural attributes of the object but also its textual semantics. While CLIP tends to retrieve icons based solely on dominant visual features (\eg, yellow-colored icons), and LLaVE retrieves less relevant results beyond the top-2, our approach consistently retrieves semantically and visually aligned icons (1st–4th results).
Moreover, our method successfully retrieves icons that reflect deeper semantic associations. For instance, the presence of bear icons in response to the ``honey'' query shows that our model understands the contextual relationship, rather than relying solely on visual resemblance. 
Also across multiple examples, our method consistently preserves both the structural attributes and the semantic meaning of the target concepts, resulting in qualitatively superior retrieval outcomes.
\subsection{Comparisons with EOL Methods} 
\label{sec:4.3}
In this section, we compare our embedding method with previous EOL-based methods~\cite{zhang2024simple, jiang2024scaling}. For PromptEOL and KEEOL, we differed the prompt setting following each method. 
Thus, we extract embeddings from each model using the prompt templates proposed by each method. 
We extract text embeddings (query) and raster image embeddings (database) from the last token's hidden state of the penultimate layer. SVG data is represented in text form, so we treat it similarly to regular text embeddings, enabling text-to-raster image retrieval using LLMs. 
Our method includes the semantic SVG module step using MLLM, while traditional EOL methods do not. For the model's generalizability, we have tested each EOL method on LLM: Mistral~\cite{jiang2023mistral7b}, MLLMs: Qwen2.5-VL-7B~\cite{bai2025qwen2}, and LLaMA-3.2-11B~\cite{grattafiori2024llama}.
\vspace{-1em}
\paragraph{Text-to-SVG Code Retrieval.} 
For text-to-SVG code retrieval, we conducted experiments on both LLM (Mistral) and MLLMs (Qwen2.5-VL-7B, LLaMA-3.2-11B).
In ~\Tref{tab:eol_ablation} (SVG), traditional EOL methods struggle to retrieve appropriate SVG code matching text queries. However, our method, which preprocesses SVG data, significantly outperforms other methods. This result implies that semantic SVG module helps LLMs and MLLMs better understand SVG data, demonstrating that our method effectively improves text-to-SVG code retrieval.
\vspace{-1em}
\paragraph{Text-to-Raster Image Retrieval.}
As shown in~\Tref{tab:eol_ablation} (Image), our method retrieves raster images more effectively compared to others. Interestingly, KEEOL performs worse than PromptEOL, suggesting that the longer query text in KEEOL might weaken its embedding representation, making it less suitable for raster image retrieval tasks.
\vspace{-1em}
\paragraph{Text-to-Raster Image Retrieval with generated SVG Code.}
\Tref{tab:eol_ablation} (Image + SVG) shows that combining raster images and generated SVG code embeddings helps to retrieve images more accurately. Thanks to the MLLM's aligned space, we can generate a combined embedding by simply feeding both raster image and SVG code to MLLM. Compared to PromptEOL and KEEOL, our method uses concise yet informative SVG data alongside raster image embeddings, capturing both visual and structural elements. Thus, our method generates embeddings that are better suited for accurate retrieval tasks.

\begin{table}[t]
    \resizebox{1\linewidth}{!}{
        \begin{tabular}{c|c|ccc}
        \toprule \midrule
        Model                       & Database Format   & Recall@1  & Recall@10  & Recall@20 \\ \midrule
        \multirow{3}{*}{LLaMA-3.2-11B}  & Image       &  \textbf{0.3387}   &  0.6526  & 0.7224 \\
                                    & Image + SVG  & 0.3110 & 0.6570 & 0.7209  \\
                                    & Image + \ensuremath{\mathrm{SVG}_{\mathrm{generated}}} &  0.3169  &   \textbf{0.6628} &   \textbf{0.7427} \\ \midrule
        \multirow{3}{*}{Qwen2.5-VL-7B} & Image       &  0.3503   &  0.6846  & 0.7456     \\
                                    & Image + SVG  &  0.3605  &  0.7122  & 0.7558  \\
                                    & Image + \ensuremath{\mathrm{SVG}_{\mathrm{generated}}} & \textbf{0.3765}  & \textbf{0.7224} & \textbf{0.7776} \\ \midrule \bottomrule
        \end{tabular}
    }
\caption{\textbf{Retrieval Performance depending on the Representation used for SVGs.} \textmd{We measured the retrieval performance for various database formats. In the database format, \textit{Image} refers to the case where only raster images are used. \textit{Image + SVG} refers to the case where raster images are used alongside the full SVG code, which does not incorporate our method. \textit{Image + $SVG_{generated}$} refers to the case where raster images are used together with the generated SVG code, which applies our method. The experimental results show that the retrieval performance improves when the generated SVG code, which incorporates our method, is used alongside the raster image.}}
\label{tab:experiments3}
\end{table}
\begin{figure}[t]
\vspace{-1em}
  \centering
  \includegraphics[width=\linewidth]{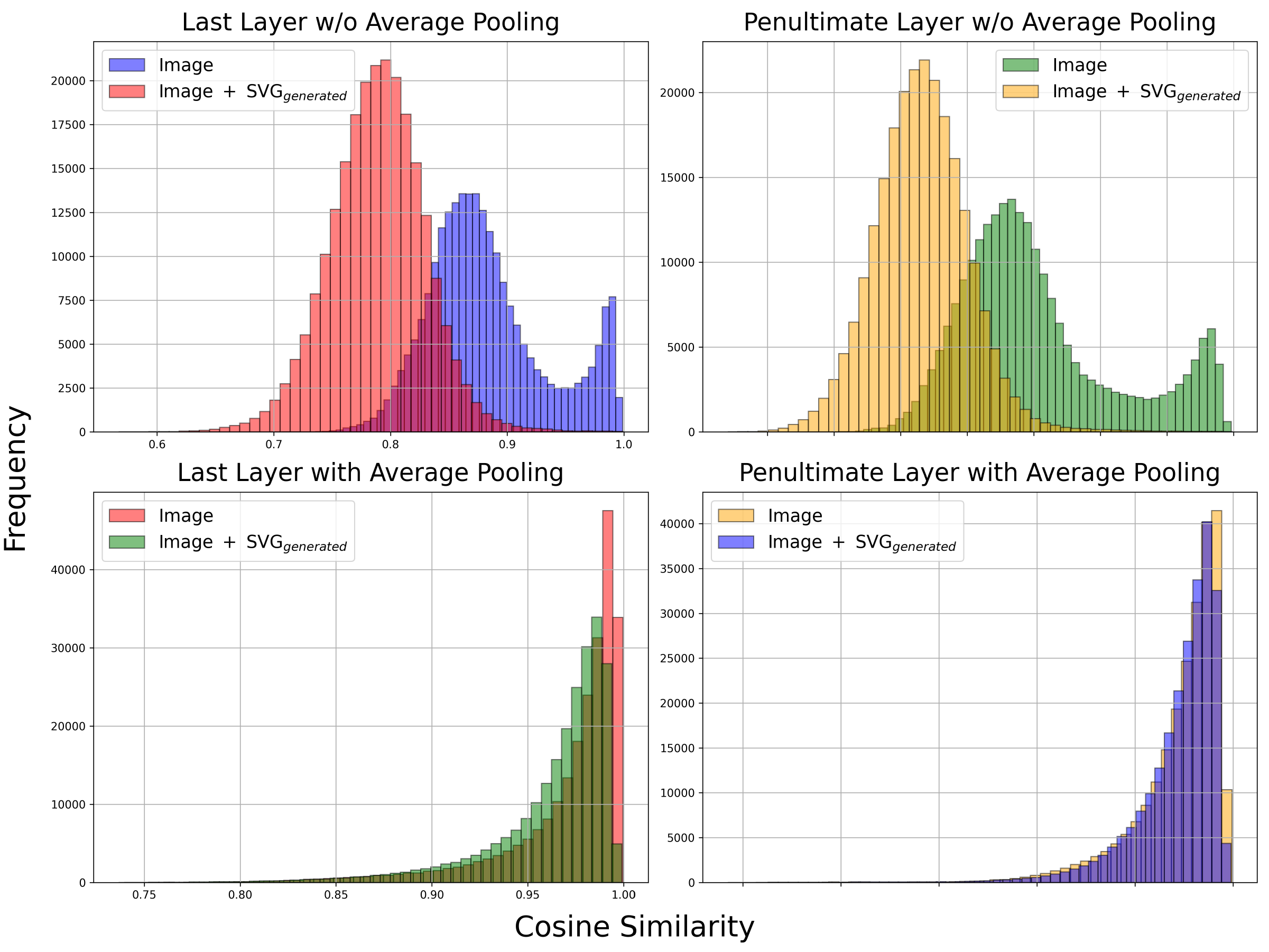}
  \caption{\textbf{Self-Cosine Similarity Distributions of Raster Image Embeddings and Embeddings Combining Raster Images with generated SVG Code in Qwen2.5-VL-7B.} \textmd{We find that the embeddings combining raster images and generated SVG code tended to have lower cosine similarity scores in all settings. This suggests that adding generated SVG code can lead to more distinct and useful embeddings.}}
  \label{fig:combine_cos_sim}
  \vspace{-1em}
\end{figure}
\subsection{Ablation Studies}
\label{sec:4.4}
\paragraph{Ablation of Semantic SVG Module.}
\Tref{tab:experiments3} compares three ways of representing SVGs: using only raster image embeddings, naively mixing raster images with raw SVG code, and mixing raster images with our semantically rewritten SVGs. A naive mixture (Image+raw SVG) lowers performance, suggesting that unprocessed SVG code introduces noise. In contrast, combining raster images with our rewritten SVGs consistently outperforms the image-only baseline. The regenerated SVG code adds clear, high-level structural cues that complement the visual signal, allowing the model to align textual and visual semantics more effectively.

Figure~\ref{fig:combine_cos_sim} further supports this observation. When SVG information is added, the resulting embeddings spread to lower cosine similarity values compared to using raster images alone. This wider distribution suggests that the rewritten SVG code injects distinctive semantic cues that help separate examples more clearly in the embedding space. By combining structural signals from SVGs with visual features from images, the model forms more discriminative multimodal representations, improving retrieval performance.
\vspace{-3mm}
\paragraph{Ablation of mEOL Prompt Formulation.}
Our formulated prompts instruct the model to summarize the input as compactly as possible. \Tref{tab:prompt_design} compares our one-word design with longer compressed outputs, ranging from two words to full sentences. Retrieval accuracy decreases as the summaries become longer. A likely reason is that longer outputs spread the salient information across multiple tokens, reducing the effectiveness of a single, coherent semantic anchor for retrieval.
\begin{table}[t]
\centering
    \renewcommand{\arraystretch}{1} 
    \setlength{\tabcolsep}{5pt} 
    \centering
    \resizebox{\linewidth}{!}
    {
        \begin{tabular}{c|cccc}
        \toprule \midrule
        Method          & Recall@1 & Recall@5 & Recall@10 & Recall@20 \\ \midrule
        One word (Ours) & \textbf{0.3169}   & \textbf{0.5596}   & \textbf{0.6628}    & \textbf{0.7427}    \\
        Two words       & 0.2340   & 0.4491   & 0.5262    & 0.6177    \\
        Three words     & 0.2384   & 0.4273   & 0.4971    & 0.5988    \\
        Four words      & 0.2253   & 0.4273   & 0.5160    & 0.5828    \\
        Sentence  & 0.1105   & 0.2137   & 0.2674    & 0.3401    \\ \midrule \bottomrule
        \end{tabular}
    }
\caption{\textbf{Ablation of mEOL prompt formulation.} \textmd{Our mEOL prompt design  represents information compactly as an one word. When prompt design request longer compressed forms, retrieval performance tends to decrease as length increases.}}
\label{tab:prompt_design}
\end{table}

\begin{table}[t]
\centering
    \renewcommand{\arraystretch}{1} 
    \setlength{\tabcolsep}{5pt} 
    \centering
    \resizebox{\linewidth}{!}
    {
        \begin{tabular}{c|cc|cc}
        \toprule \midrule
        Model & Penultimate Layer & Last token & Recall@1 & Recall@5 \\ \midrule

        \multirow{4}{*}{LLaMA-3.2-11B} & \textcolor{red}{\ding{55}} & \textcolor{red}{\ding{55}} & 0.0262 & 0.0610 \\ 
                              & \textcolor[HTML]{228B22}{\ding{51}} & \textcolor{red}{\ding{55}} & 0.0044 & 0.0145 \\
                              & \textcolor{red}{\ding{55}} & \textcolor[HTML]{228B22}{\ding{51}} & 0.3212 & 0.5698 \\ 
                              & \textcolor[HTML]{228B22}{\ding{51}} & \textcolor[HTML]{228B22}{\ding{51}} & \textbf{0.3387} & \textbf{0.5785} \\  \midrule
        \multirow{4}{*}{Qwen2.5-VL-7B} & \textcolor{red}{\ding{55}} & \textcolor{red}{\ding{55}} & 0.0073 & 0.0203 \\ 
                              & \textcolor[HTML]{228B22}{\ding{51}} & \textcolor{red}{\ding{55}} & 0.0015 & 0.0087 \\
                              & \textcolor{red}{\ding{55}} & \textcolor[HTML]{228B22}{\ding{51}} & 0.3387 & 0.5930 \\ 
                              & \textcolor[HTML]{228B22}{\ding{51}} & \textcolor[HTML]{228B22}{\ding{51}} & \textbf{0.3503} & \textbf{0.6177} \\ \midrule \bottomrule
        \end{tabular}
    }
\caption{\textbf{Ablation on Raster Image Embedding Layer and Pooling Method.} \textmd{Using the last token from the penultimate layer without pooling achieves the best performance. The last token indicates that average pooling has not been applied.}}
\label{tab:embedding_exp5}
\end{table}
\paragraph{Ablation of Token Embedding for Raster Images.}
Prior EOL methods~\cite{jiang2024scaling, thirukovalluru2024geneol, zhang2024simple, lei2024meta} typically uses the method to extract the sentence's representation from either the averaging all input token embeddings or the last token of the last hidden layer.
As we extend EOL methods to Multimodal Large Language Models (MLLMs), we need to verify whether the tendencies of MLLMs regarding these options are similar to those observed in traditional EOL methods. 
\Tref{tab:embedding_exp5} represents the result of the embedding methods and the use of the penultimate layer. As shown, using the penultimate layer with the simple last token shows the best results, followed by the use of the last layer with the last token.
\begin{figure}[t]
  \centering
  \includegraphics[width=\linewidth]{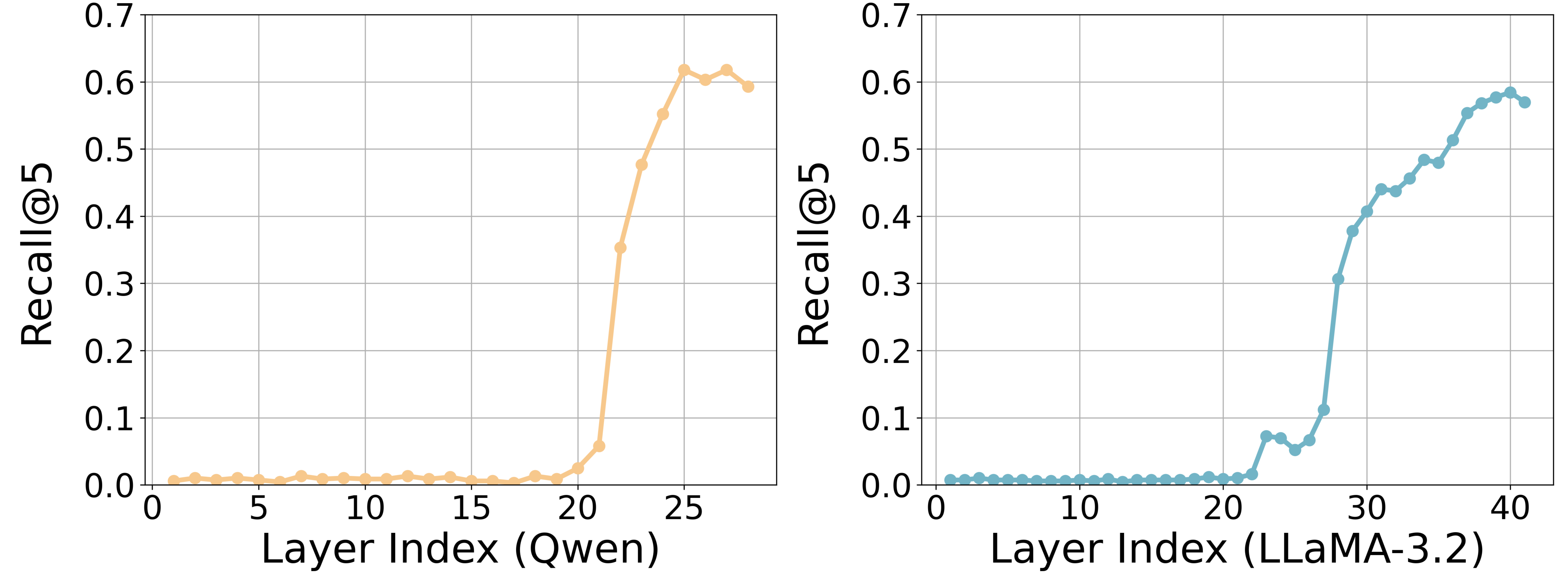}
  \caption{\textbf{Ablation on Layer Index for Raster Image Embedding Extraction. \textmd{We evaluate retrieval performance across layers when selecting the last token embedding. Performance generally improves in the later layers, with the penultimate layer consistently offering the best trade-off between stability and retrieval quality.}
}}
  \label{fig:layer_wise}
\end{figure}
\begin{figure}[t]
\vspace{-1em}
  \centering
  \includegraphics[width=\linewidth]{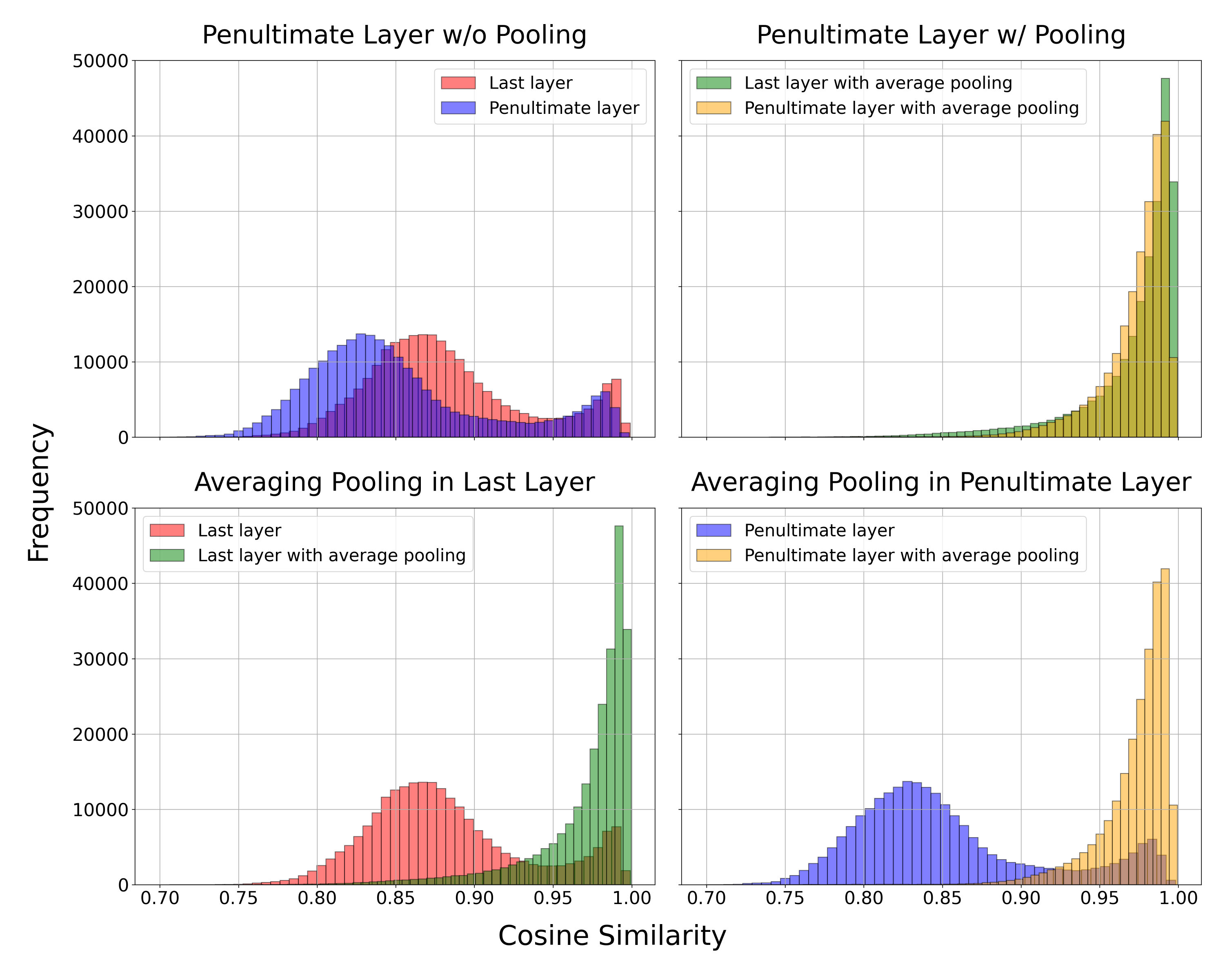}
  \caption{\textbf{Self-Cosine Similarity Distributions of Raster Image Embeddings in Qwen2.5-VL-7B, evaluated across layers and pooling strategies.}
  \textmd{\textbf{(top)} Self-similarity distributions when extracting the last-token embedding from different layers. Using the penultimate layer shifts the distribution toward lower cosine similarity values compared to the last layer, indicating more diverse and expressive representations. 
  \textbf{(bottom)} 
  Self-similarity distributions under different embedding extraction methods. Embeddings obtained without average pooling show substantially lower similarity, while pooling collapses the distribution toward higher values. Taken together, the results show that the penultimate layer without pooling yields the most discriminative embedding space.
  }}
\vspace{-1em}
  \label{fig:cos_sim}
\end{figure}
\vspace{-1em}
\paragraph{Ablation of Layer Index.}
Specifically, we examine which hidden layers of the MLLM are most suitable for obtaining useful embeddings. Figure~\ref{fig:layer_wise} indicates that recall@5 sharply increase in the later layers for both models, \ie, Qwen2.5-VL-7B, LLaMA-3.2-11B. This implies that embeddings from later layers are more informative compared to the early layers. Therefore, following traditional EOL methods, we compare embeddings from the last layer and the penultimate layer. 
Figure \ref{fig:cos_sim} (top) shows the self-similarity distribution of raster image embeddings extracted for the penultimate layer and last layer. The left plot shows the distribution when using the last token embedding, and the right plot shows the distribution with average pooling over all tokens. In both cases, embeddings from the last layer exhibit a right-shifted distribution compared to those from the penultimate layer, indicating lower variance and expressiveness of the embeddings, similar results are shown at~\cite{ethayarajh2019contextual}. Thus, using the penultimate layer can improve embedding expressiveness.
\paragraph{Ablation of Embedding Method.}
Next, we test the effect of two embedding methods: using the last token, \ie, newly generated one, or average pooling the whole tokens' hidden vectors. Figure~\ref{fig:cos_sim} (bottom) demonstrates that when average pooling is used, the self-similarity distribution becomes concentrated near 1. This means  pooling the whole hidden states makes embeddings overly similar, reducing their distinct semantic information for retrieval tasks.
In the context of SVGs, where each sample may contain many components with diverse structural formats and numerical values, simple averaging over all token embeddings may fail to capture the proper semantic or textual information of the input. But the embedding derived from the last token, which implies the MLLM's interpretation of the input tokens, can be more effective in preserving semantic or textual information.
Therefore, we conclude that using the last token embedding is better suited for our framework, which handles the structured and variable nature of SVG inputs.

\section{Conclusion}
We presented mEOL, a training-free instruction-guided multimodal embedding framework that maps text, raster images, and SVG code into an aligned representation space using a MLLM. By extending EOL to multimodal inputs and designing modality-specific instructions, mEOL compresses diverse inputs into compact, semantically meaningful embeddings without any additional training or fine-tuning. Combined with a semantic SVG rewriting module that assigns interpretable identifiers and simplifies SVG structure through visual reasoning, our approach makes the symbolic organization of vector graphics directly accessible to a retrieval system. On a repurposed VGBench, we instantiated, to our knowledge, the first text-to-SVG retrieval benchmark and evaluated text-to-image and text-to-image+SVG retrieval on the same icon corpus. These results suggest that prompt-level control is a viable alternative to parameter-level training for structure-aware retrieval, and point toward extending instruction-guided, training-free embeddings to broader vector graphics domains and other structured modalities. While our one-word formulation emerged as the most effective and stable design for unifying heterogeneous modalities, future work may explore more fine-grained or attribute-aware instruction strategies to extend mEOL toward richer forms of multimodal control.
\section*{Acknowledgment}
This work was supported by Institute of Information \& Communications Technology Planning \& Evaluation (IITP) grant funded by the Korea government(MSIT) (No.RS-2023-00225630, Development of Artificial Intelligence for Text-based 3D Movie Generation). T.-H. Oh was partially supported by the ‘Ministry of Science and ICT’ and NIPA ("HPC Support" Project), KAIST Cross-Generation Collaborative Lab Project.
{
    \small
    \bibliographystyle{ieeenat_fullname}
    \bibliography{main}
}

\end{document}